\documentclass{article}
\usepackage{spconf,amsmath,graphicx}
\usepackage{booktabs}
\usepackage{amssymb}
\usepackage{bm}
\usepackage{makecell}

\usepackage{enumitem}
\setlist{nosep, leftmargin=14pt}

\usepackage{mwe} 
\usepackage{multicol}
\usepackage{multirow}

\title{ Breast Lesion Diagnosis Using Static Images and Dynamic Video }
\name{Yunwen Huang $^{1}$\sthanks{Equal contribution.} \qquad Hongyu Hu$^{1}$ \footnotemark[1]  \qquad Ying Zhu$^{2}$ \qquad Yi Xu$^{1}$\sthanks{Corresponding author.}}

\address{$^{1}$ Shanghai Jiao Tong University \\
    $^{2}$ Department of Ultrasound, Ruijin Hospital, Shanghai Jiaotong University School of Medicine }
\begin{document}
%
\maketitle

\section*{Abstract}

Deep learning based Computer Aided Diagnosis (CAD) systems have been developed to treat breast ultrasound. Most of them focus on a single ultrasound imaging modality, either using representative static images or the dynamic video of a real-time scan. In fact, these two image modalities are complementary for lesion diagnosis. Dynamic videos provide detailed three-dimensional information about the lesion, while static images capture the typical sections of the lesion. In this work, we propose a multi-modality breast tumor diagnosis model to imitate the diagnosing process of radiologists, which learns the features of both static images and dynamic video and explores the potential relationship between the two modalities. Considering that static images are carefully selected by professional radiologists, we propose to aggregate dynamic video features under the guidance of domain knowledge from static images before fusing multi-modality features. Our work is validated on a breast ultrasound dataset composed of 897 sets of ultrasound images and videos. Experimental results show that our model boosts the performance of Benign/Malignant classification, achieving $90.0\%$ in AUC and $81.7\%$ in accuracy.

\textit{\textbf{Index Terms—}} Multi-Modal Fusion, Breast Cancer Diagnosis, Ultrasound Imaging

\section{Introduction}

Breast cancer is one of the most prevalent cancer in the world. With the development of modern medical technology, breast cancer treatment is highly effective if identified early. Breast ultrasound (BUS) is a frequently utilized imaging modality for diagnosing breast cancer, as it is convenient, non-invasive, and radiation-free.
Generally, a BUS exam requires a radiologist to select representative static images that depict the maximum diameter and depth of the lesion or highlight typical malignant characteristics and record a real-time scan of the lesion as a dynamic video. 
Deep learning based computer-aided diagnosis (CAD) systems have been developed to reduce the burden of radiologists and perform a more efficient and precise diagnosis.

Current deep learning based breast ultrasound CAD systems aim to incorporate domain knowledge into the feature extraction process of ultrasound imaging, which can be summarized into two categories.
The first type uses static images as the input modality. With the success of ResNet \cite{he_deep_2015} and Vision Transformer \cite{dosovitskiy_image_2021} in image processing, they have been adopted as the backbone network for numerous studies~\cite{BMVC}. 
For example, based on ResNet, gaze tracking data of the radiologists \cite{droste_ultrasound_2019} is utilized to model the attention radiologists pay to images. As breast lesion mainly consists of four layers: the subcutaneous fat layer, gland layer, muscle layer, and thorax layer, HoVer-Transformer \cite{mo_hover-trans_2022} introduces anatomy-aware information by extracting inter-layer and intra-layer spatial information. 
Multi-task frameworks are proposed in \cite{singh_efficient_2019, yu_computer-aided_2020} as tumor segmentation can provide knowledge on the size and shape of the tumor, which is essential to enhance the performance for breast tumor diagnosis.
The second type uses dynamic videos. Since videos are sequences of images, 3DResNet \cite{res3d} and TimeSformer \cite{bertasius_is_2021} have been proposed to learn the temporal information. Volume reconstruction is proposed in \cite{luo_self_2021} to extract the three-dimensional feature of the lesion contained in the video.
Self-supervised learning is utilized in \cite{chen_uscl_2021} to learn compact semantic clusters from dynamic videos.
Keyframes extraction is performed in \cite{huang_extracting_2022, wang_key-frame_2022} and is further utilized to guide video feature extraction in \cite{wang_key-frame_2022}. 
Furthermore, temporal attention is guided by the brightness of the video frame as radiologists' attention roughly matches the brightness curve as indicated in \cite{chen_domain_2021}.
Though the above methods consider the characteristics of each modality and embed domain knowledge, features learned from a single modality are incomplete, as static images are sections selected by radiologists according to medical knowledge, while dynamic videos provide complete information about the lesion and may avoid section discrepancy according to the study in \cite{yang_section_2022}. 

To address this issue, we propose a breast lesion diagnosis framework using static images and dynamic video.
Our model mimics the radiologist's workflow by processing the dynamic video to form a general lesion representation and integrating it with static images to obtain detailed lesion characteristics. 
Significantly, domain knowledge of typical malignant characteristics, size and shape of the lesion is encoded in the static images, which provides promising guidance for learning features in dynamic video via a frame-wise attention mechanism.
The contribution of our work is three-fold:
\begin{itemize}
    
    \item We propose a breast ultrasound tumor diagnosis framework that processes multi-modality features from multiple static images and a single dynamic video, which matches the real-world diagnosis process.
    \item An image-guided video feature enhancement and aggregation module is proposed to embed domain knowledge from static images to dynamic video.
    \item Extensive experiments on a real-world multi-modality breast cancer dataset are conducted to validate the effectiveness of our method, showing superior performance over baseline models.
\end{itemize}

\section{Method}

\subsection{Overview}

\begin{figure}[]
\begin{minipage}[b]{1.0\linewidth}
  \centering
  \centerline{\includegraphics[width=1\linewidth]{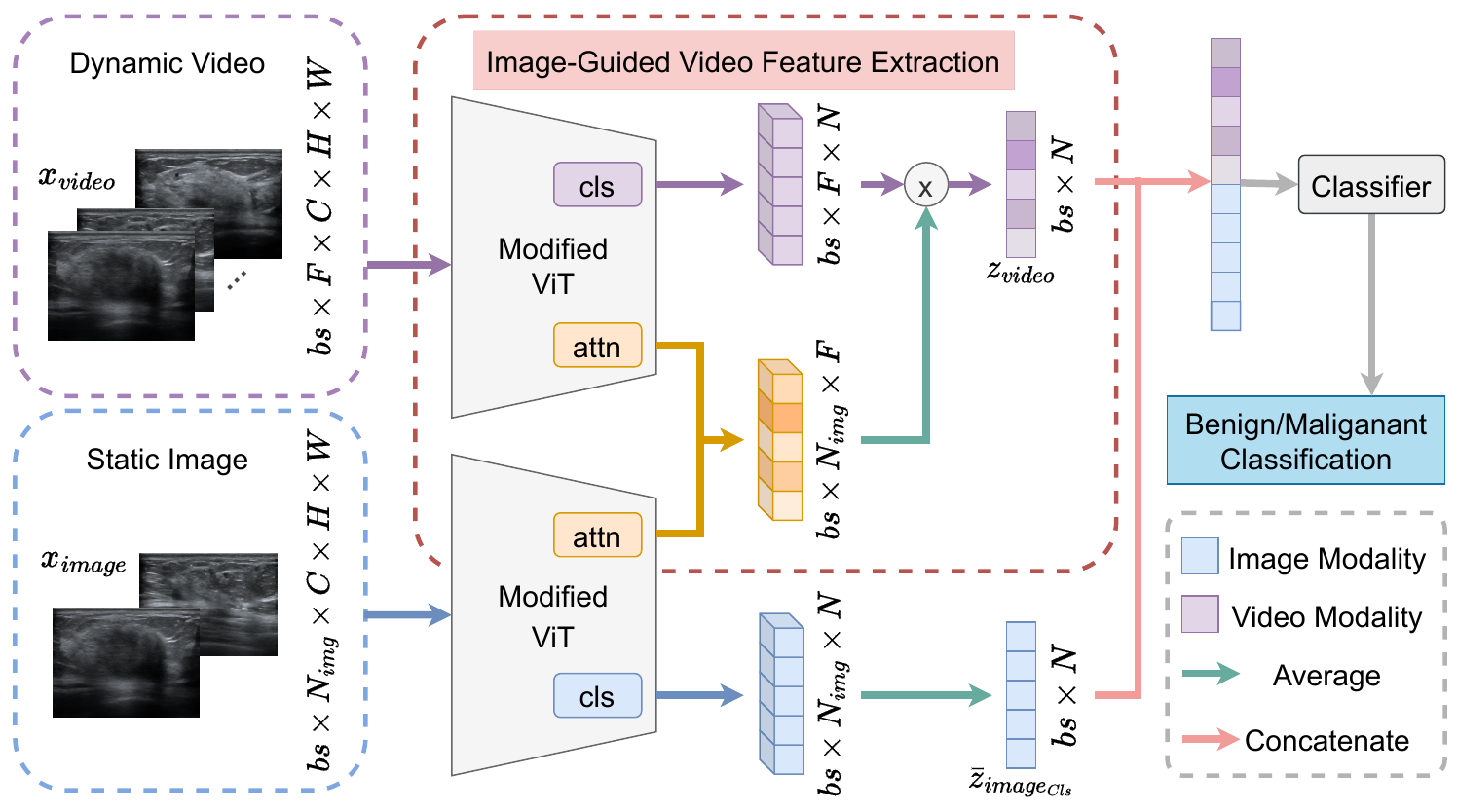}}
\end{minipage}\caption{Overview of the proposed multi-modality breast tumor diagnosis model. A modified vision transformer is utilized to extract general feature representation and the radiologist's attention feature. Static images are further referred to for video feature aggregation, and features extracted from different modalities are fused to form a comprehensive lesion feature, which is fed into a classification head.}
\label{fig:framework}
\end{figure}

The proposed multi-modality breast tumor diagnosis model aims to extract comprehensive features from static images and dynamic video. 
The most distinguishable characteristics of the lesion could be easily learned from the static images, which consist of typical sections selected by professional radiologists. Meanwhile, dynamic videos provide more general information about the tumor. It is rational to extract video features under the guidance of image modality and then fuse the features from two modalities for comprehensive representation.
The overall framework is shown in Fig.\ref{fig:framework} and is divided into two stages: Image-guided video feature extraction and multi-modality fusion.


\subsection{Image-Guided Video Feature Extraction}
\subsubsection{Modified Vision Transformer}

\begin{figure}[]
\begin{minipage}[b]{1\linewidth}
  \centering
  \centerline{\includegraphics[width=0.9\linewidth]{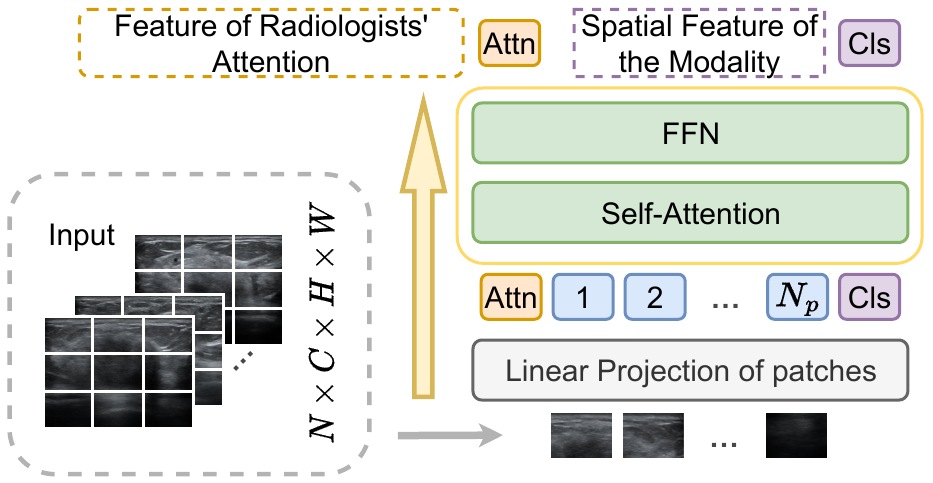}}
\end{minipage}
\caption{Illustration of the Modified Vision Transformer.}
\label{fig:vit}
\end{figure}

To learn the general characteristics and domain knowledge guided lesion attention of each modality, we propose a modified vision transformer backbone, illustrated in Fig.\ref{fig:vit}. 
For every single image or video frame, it is reshaped into $N_p$ patch sequences, where $N_p = H\times W / P^2$ and $P\times P$ is the resolution of the image patch. 
Apart from patch tokens, we introduce classification token \verb=[cls]= and attention token \verb=[Attn]= to learn the general representation of the input modality and the radiologist's attention separately. The attention token  \verb=[Attn]= is further utilized to capture the inherent relationships between image and video modalities.
For the input batch ($N$ ultrasound images or video frames) $\bm{x}\in \mathbb{R}^{N \times C \times H \times W}$, the output is $\bm{z_{Cls}} \in \mathbb{R}^{N\times d_{out}}$ and $\bm{z_{Attn}} \in \mathbb{R}^{N\times d_{out}}$. In this work, the output dimension is $d_{out} = 1000$.


\subsubsection{Image-Guided Video Feature Aggregation } \label{attn}

As static images are selected by radiologists and contain domain knowledge like typical malignant lesion characteristics, we propose to utilize image modality as a solid reference to guide the feature aggregation of video frames.
Specifically, cross-attention is adopted to obtain frame-wise attention based on static images and aggregate feature representation of the video modality. 
A special \verb=[Attn]= token is utilized in both modalities to aggregate domain-knowledge-based attention information.
Attention token of image modality $\bm{z_{image_{Attn}}}$ serves as the query for strong reference, the attention token of video frames $\bm{z_{video_{Attn}}}$ are keys, and the classification tokens of video frame $\bm{z_{video_{Cls}}}$ are values in cross-attention.
Generally, this cross-attention operation is formulated in Eq.\ref{cross-attention}.

\begin{equation}\label{cross-attention}
\bm{z_{video}} = \text{softmax}(\bm{\bar z_{image_{Attn}}}\bm{z_{video_{Attn}}^T})\bm{z_{video_{Cls}}}
\end{equation}

Note that if there are multiple static images selected for a single patient, their contributions to the frame-wise attention are averaged.
The proposed attention module can pay more attention to the keyframes selected by the model, imitating the temporal attention of radiologists and the diagnosis process in which radiologists assess the entire lesion using dynamic video and focus on keyframes to make a thorough diagnosis.

\subsection{Multi-modality Fusion}


In practice, radiologists deal with breast ultrasound imaging by generally diagnosing the lesion based on the dynamic video and merging the static images to obtain a more accurate diagnosis. 
The dynamic video features $\bm{z_{video}}$ provide an overall representation of the lesion, while the static image features $\bm{z_{image_{Cls}}}$ are more targeted to depict the tumor. Features from both modalities are complementary, and we concatenate the video and image features to form a comprehensive feature representation of the lesion, similar to the procedure how radiologists diagnose. A classification head is further added to conduct downstream diagnosis task (malignancy prediction) and give final diagnosis $\bm{\widehat{y}}$, as formulated in Eq.\ref{cls}. 
\begin{equation}\label{cls}
\bm{\widehat{y}} = \text{head}(\text{Concat}(\bm{z_{video}}, \bm{\bar z_{image_{Cls}}}))
\end{equation}

Let $\bm{\theta}$ represent the trainable parameters of the model.
The network with parameter $\bm{\theta}$ is optimized by cross-entropy loss. Given the ground truth $\bm{y}$ and predicted output probability $\bm{\widehat{y}}$, the optimization goal of our model is formulated in Eq.\ref{optimize}

\begin{equation}\label{optimize}
\mathop{\arg\min}_{\bm{\theta}} Loss=-[\bm{y} \log \bm{\widehat{y}}+(1-\bm{y}) \log (1-\bm{\widehat{y}})]
\end{equation}

\section{Experiments}

\subsection{Experimental Settings}
\textbf{Dataset.} Our model is validated on a multi-modality ultrasound dataset with 897 sets of ultrasound images and videos, where 508 patients have malignant tumors, and 389 patients have benign tumors. A single set of data contains several static images and a dynamic video. The static images are the representative sections chosen by the radiologists, and the dynamic video is the corresponding real-time scan of the lesion.\\
\textbf{Evaluation Protocol.} To evaluate the performance of our model, the k-fold (k=5) cross-validation method is applied. Area under the ROC curve (AUC), accuracy (Acc), and F1-score are used as evaluation metrics.\\
\textbf{Implementation Details.}
ViT-Tiny \cite{touvron_training_2021} is adopted as the backbone network. For data augmentation, all images are resized to 224 × 224 pixels, followed by random flipping and color jittering to prevent overfitting. All videos are randomly sampled to get 16 video frames, and each video frame follows the same data augmentation steps as images. Adam optimizer \cite{adam} is adopted to train all models with an initial learning rate of $1e-5$ for 50 epochs. All experiments were conducted on NVIDIA TITAN X GPU and NVIDIA RTX 3090 GPU. 

\subsection{Ablation Study}
\begin{table}[]
 \caption{Comparison of model settings using ViT-Tiny as the backbone. Benign/malignant classification performance (mean±std\%) is reported.}
\resizebox{1\linewidth}{!}{

\begin{tabular}{ccccc}
\hline
 \makecell{Feature\\ Modality} & Attn & AUC & F1 Score & Acc \\ \hline

Video                                                                      & {[}cls{]}  & $82.9 \pm 2.5$       & $69.6 \pm 8.7$      & $71.7 \pm 5.7$      \\
Video                                                                      & {[}Attn{]} & $87.8 \pm 2.1$       & $82.1 \pm 1.2$      & $80.4 \pm 1.4$      \\
Multi                                                                      & -          & $ 87.4 \pm 3.4$      & $78.2 \pm 6.3$      & $77.9 \pm 5.3$      \\
Multi                                                                      & {[}cls{]}  & $87.8 \pm 3.3$       & $79.4 \pm 4.8$      & $78.5 \pm 4.5$      \\ \hline
\textbf{Multi}                                                             & {[}Attn{]} & $ \bm{90.0 \pm 1.4}$ & $\bm{82.6 \pm 0.9}$ & $\bm{81.7 \pm 0.9}$ \\ \hline
\end{tabular}
}
\label{tab:ablation}
\end{table}

Ablation studies are conducted to validate the effectiveness of our image-guided video feature extraction module and multi-modality fusion module. Generally, we study the feature modality to feed into the downstream classifier and how static images should guide feature representation of videos in cross-attention (Attn denotes which token to use in cross-attention.). Ablation results are shown in Table \ref{tab:ablation}.




\textbf{Effectiveness of} \verb=[Attn]=. Table.\ref{tab:ablation} also shows that under the setting of multi-modality input, with the fixed feature modality, applying extra attention token \verb=[Attn]= in cross-attention performs better than classification token \verb=[cls]=(87.8\% vs. 82.9\%, 90.0\% vs. 87.8\% in AUC). The classification token is directly fed into the downstream classifier and thus focuses on representing distinguishable features, ignoring the radiologists' attention during diagnosis. Therefore, it is necessary to add an extra \verb=[Attn]= in the modified transformer as the strong domain knowledge to mimic radiologists' frame-wise attention.


\textbf{Effectiveness of Multi-Modality Fusion.} It is also displayed in Table.\ref{tab:ablation} that with the fixed token to use in cross-attention, utilizing the multi-modality feature achieves better performance over the video feature (87.8\% vs. 82.9\%, 90.0\% vs. 87.8\% in AUC). Though static images guide the feature extraction of dynamic video, feature representations of video modality could not capture complementary information embedded in image modality. Therefore, it is essential to fuse two modalities for a more comprehensive feature representation. 

\subsection{Comparison with Baseline Models}
\begin{table}[]
    \centering
    \caption{Comparison of benign/malignant classification performance (mean±std\%) on different methods with single or multi modality used.}
    
    \resizebox{1\linewidth}{!}{
    \begin{tabular}{ccccc}
    \toprule
    Modality & Method &  AUC & F1 Score & Acc \\
    \midrule
    Image & ResNet18\cite{he_deep_2015} & $84.6 \pm 2.6$       & $77.9 \pm 3.7$      & $75.3 \pm 3.5$      \\
    Image& ViT-Tiny\cite{dosovitskiy_image_2021} & $87.3 \pm 1.5$       & $77.0 \pm 4.7$      & $76.9 \pm 3.4$      \\
    Video & USCL\cite{chen_uscl_2021} & $85.6 \pm 1.3$       &  $73.9\pm 9.9$      & $73.1 \pm 6.1$      \\
    Video & R3D18\cite{res3d} & $77.9 \pm 1.0$       &  $68.6\pm 11.9$      & $68.4 \pm 5.5$      \\
    Video & TimeSformer\cite{bertasius_is_2021} & $87.3 \pm 0.9$       &  $74.4\pm 6.5$      & $75.0 \pm 3.7$      \\
    \midrule
    \multicolumn{1}{c}{\multirow{5}{*}{Multi}} & Resnet18 \cite{he_deep_2015}  &$ 85.2 \pm 3.6$ & $ 75.9 \pm 8.9$ &$ 74.6 \pm 4.9$\\
    & ViT-Tiny \cite{dosovitskiy_image_2021}  & $ 87.4 \pm 3.4$      & $78.2 \pm 6.3$      & $77.9 \pm 5.3$\\
    & USCL \cite{chen_uscl_2021}  &$ 85.0 \pm 0.9$ &$ 76.1 \pm 7.3$ &$ 74.9 \pm 4.3$\\
    & R3D18 \cite{res3d} &   $ 80.7 \pm 2.0$& $ 72.6 \pm 5.5$&$ 71.8 \pm 3.5$\\
    & TimeSformer \cite{bertasius_is_2021}  & $ 88.5 \pm 1.1$& $ 80.9 \pm 7.2$&$ 80.5 \pm 4.5$\\
    \midrule
    Multi & Ours  & $ \bm{90.0 \pm 1.4}$ & $\bm{82.6 \pm 0.9}$ & $\bm{81.7 \pm 0.9}$ \\
    \bottomrule
    \end{tabular}
    }
    \label{tab:comparison}
\end{table}

To further evaluate the effectiveness of the proposed method, we compare it with baseline models designed for different modalities.\\
\textbf{Baseline Models.} ResNet\cite{he_deep_2015} and ViT~\cite{dosovitskiy_image_2021} are two commonly-used backbones for visual tasks. 
USCL~\cite{chen_uscl_2021} is a self-supervised ultrasound image diagnosis model that generates sample pairs from ultrasound videos to force semantic cohesion appearing at the volume level. 
R3D18 ~\cite{res3d} expands convolution into 3D convolution to deal with video classification tasks.
TimeSformer~\cite{bertasius_is_2021} is a SOTA video classification method that performs temporal and spatial attention separately within a block.
The methods above cover competitive methods designed for image and video feature extraction problems and those designed for ultrasound imaging.

Note that the above-mentioned baselines all work on single modality. For fair comparison, we extend them to multi-modality input.



\textbf{Experimental Results.}Baseline models trained on a single modality are shown in the first group of experiments in Table \ref{tab:comparison}.
Resnet18 and ViT-Tiny do not perform well since only spatial features of the static images are extracted, and section discrepancy may emerge as images lack information on the tumor's surroundings.
USCL performs poorly because equal attention is paid to each frame, whereas our model concentrates on keyframes like radiologists.
R3D18 achieves the worst result as the extracted three-dimensional feature of dynamic video is too general and lacks tumor-aware local information. 
TimeSformer achieves relatively better performance, which can be attributed to the temporal attention \cite{bertasius_is_2021} mechanism that learns to focus on typical sections. However, the model size of TimeSformer is 116.1M, and our model is 12.3M since frame-wise attention is explicitly incorporated according to static images in our model, reducing temporal attention computation and introducing medical bias.  

As the above methods only consider the single modality of breast ultrasound imaging, which is insufficient, we modify the baseline models and perform experiments on multi-modality input in the second group of Table \ref{tab:comparison}. With complementary information introduced, most baseline models show better performance with multi-modality input than with single-modality input, illustrating the importance of combining local and global information.
Despite the performance boost, our work still outperforms the selected methods in the Benign/Malignant classification task, achieving an AUC of $90.0\%$, an F1 score of $82.6\%$ and an accuracy of $81.7\%$, which can be attributed to aggregating dynamic video features under the guidance of domain knowledge and further forming a comprehensive feature representation based on multi-modality inputs.

\subsection{Visualization of Attention Maps}
\begin{figure}[]
\begin{minipage}[b]{1\linewidth}
  \centering
  \centerline{\includegraphics[width=0.7\linewidth]{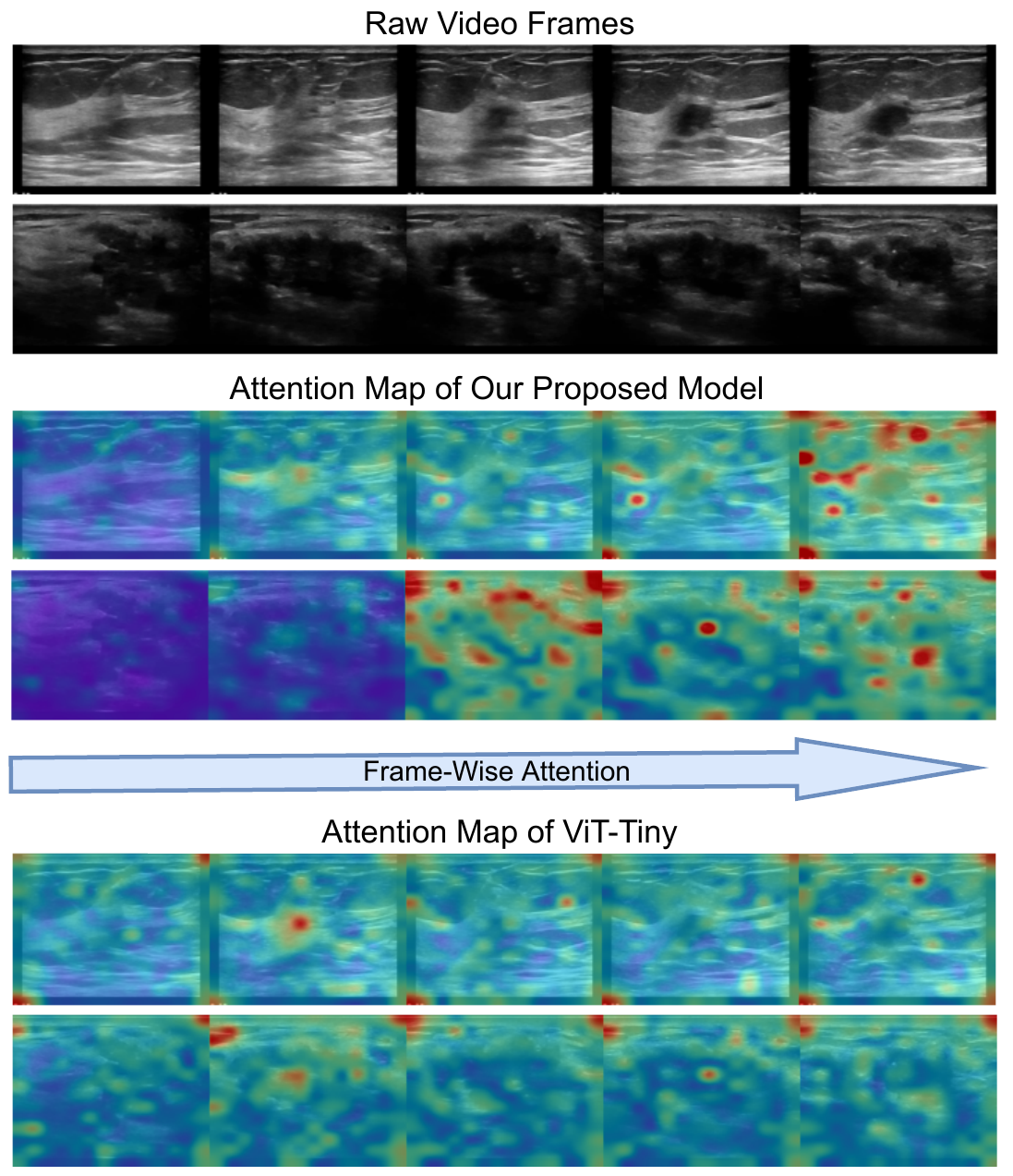}}
\end{minipage}
\caption{Visualization of attention maps produced by our model and ViT-Tiny.}
\label{fig:vis}
\end{figure}

To further illustrate the effectiveness of our model, we visualize the attention maps of breast ultrasound video produced by our multi-modality model and ViT-Tiny in Fig.\ref{fig:vis}. 

Fig.\ref{fig:vis} shows that our proposed model can capture the characteristics of the tumor as more attention is paid to the tumor area within a frame. 
Furthermore, the sequence of video frames is rearranged by frame-wise attention scores for better visualization in Fig.\ref{fig:vis} (The further to the right, the more attention the frame is paid). It can be found that ordinary ViT-Tiny pays equation attention to every frame, while our model concentrates on keyframes that capture the tumor more clearly. Generally, visualization results indicate that our proposed method can capture both local spatial attention and frame-wise importance.



\section{Conclusion}

Radiologists diagnose breast tumors by jointly considering static images and a dynamic video of a real-time scan. In this paper, we follow the diagnosing process of radiologists and propose a multi-modality breast tumor diagnosis model. To make good use of domain knowledge, an ultrasound image-guided feature extraction module is designed to aggregate the feature of dynamic video, and the features from the two modalities are further fused.
Experimental results demonstrate the effectiveness of our proposed modules, and our model outperforms other models in the breast tumor classification problem.
Our work suggests that integrating multi-modality information and adding medical inductive bias improve the performance. We hope this attempt can improve the construction of deep learning based computer-aided diagnosis systems.

\textbf{Compliance with Ethical Standards}: All procedures performed in studies involving human participants were in accordance with the ethical standards of the institutional and/or national research committee.

\textbf{Acknowledgment}: This work was supported in part by National Natural Science Foundation of China (61671298), 111 project (BP0719010), Shanghai Science and Technology Committee (18DZ2270700) and Shanghai Jiao Tong University Science and Technology Innovation Special Fund (ZH2018ZDA17).


\bibliographystyle{IEEEbib}

\end{document}